%% file: 2019-kitti-vlp32.tex
\documentclass[a4paper, 10pt, conference]{ieeeconf}      % Use this line for a4 paper

\IEEEoverridecommandlockouts                              % This command is only needed if 
\newcommand{\eqpunct}[1]{\:\text{#1}}

\usepackage[pdftex]{graphicx}

\usepackage{import} % relative path for input

\usepackage{amssymb} % also loads amsfonts

\usepackage{bm} % bold math

\usepackage{mathtools}

\usepackage{xcolor}

\usepackage{siunitx}

\usepackage[capitalize]{cleveref}
\crefname{equation}{}{}

\usepackage{import}

\usepackage{booktabs}

\usepackage{tikz}

\newcommand\copyrighttext{%
	\footnotesize \textcopyright\ 2019 IEEE. Personal use of this material is permitted. Permission from IEEE must be obtained for all other uses, in any current or future media, including reprinting/republishing this material for advertising or promotional purposes, creating new collective works, for resale or redistribution to servers or lists, or reuse of any copyrighted component of this work in other works. DOI: 10.1109/ITSC.2019.8917011
}%
\makeatletter
\newcommand\copyrightnotice{%
	\begin{tikzpicture}[remember picture,overlay]%
	\node[anchor=south,yshift=9pt,xshift=\dimexpr1in+\hoffset+\oddsidemargin+.5\textwidth-.5\paperwidth] at (current page.south) {\fbox{\parbox{\dimexpr\textwidth-\fboxsep-\fboxrule\relax}{\copyrighttext}}};%
	\end{tikzpicture}%
	\vspace{-10pt}%
}
\makeatother

\title{\LARGE \bf
Training a Fast Object Detector for LiDAR Range Images \\Using Labeled Data from Sensors with Higher Resolution
}

\author{Manuel Herzog$^{1}$ and Klaus Dietmayer$^{1}$% <-this % stops a space
\thanks{$^{1}$Manuel Herzog and Klaus Dietmayer are with the Institute of Measurement, Control, and Microtechnology,
Ulm University,
89081 Ulm, Germany
{\tt\small \{manuel.herzog, klaus.dietmayer\}@uni-ulm.de}}%
}

\begin{document}

\maketitle
\copyrightnotice
\thispagestyle{empty}
\pagestyle{empty}

%%%%%%%%%%%%%%%%%%%%%%%%%%%%%%%%%%%%%%%%%%%%%%%%%%%%%%%%%%%%%%%%%%%%%%%%%%%%%%%%

\begin{abstract}

In this paper, we describe a strategy for training neural networks for object detection in range images obtained from one type of LiDAR sensor using labeled data from a different type of LiDAR sensor.
Additionally, an efficient model for object detection in range images for use in self-driving cars is presented.
Currently, the highest performing algorithms for object detection from LiDAR measurements are based on neural networks.
Training these networks using supervised learning requires large annotated datasets.
Therefore, most research using neural networks for object detection from LiDAR point clouds is conducted on a very small number of publicly available datasets.
Consequently, only a small number of sensor types are used.
We use an existing annotated dataset to train a neural network that can be used with a LiDAR sensor that has a lower resolution than the one used for recording the annotated dataset.
This is done by simulating data from the lower resolution LiDAR sensor based on the higher resolution dataset.
Furthermore, improvements to models that use LiDAR range images for object detection are presented.
The results are validated using both simulated sensor data and data from an actual lower resolution sensor mounted to a research vehicle.
It is shown that the model can detect objects from \SI[detect-weight=true,detect-family=true]{360}{\degree} range images in real time.

\end{abstract}

%%%%%%%%%%%%%%%%%%%%%%%%%%%%%%%%%%%%%%%%%%%%%%%%%%%%%%%%%%%%%%%%%%%%%%%%%%%%%%%%
\section{Introduction}

Reliable detection of other road users in the environment of the vehicle is one of the challenges for self-driving cars.
Together with cameras and radars, LiDAR sensors are one of the major sensor types used for gathering information about the surrounding of the vehicle.
While LiDAR sensors do not achieve the resolution of cameras, they can provide reliable and precise distance measurements.
At the same time, LiDAR sensors typically provide a significantly higher resolution and accuracy than automotive radar sensors.

The measurements from a LiDAR sensor can be represented as a set of points (``point cloud") where each point contains information about its three-dimensional position.
In addition to the position data, LiDAR sensors can also provide information about the reflectivity of the surface.
There is already a large amount of literature on detecting objects from point clouds which are provided by LiDAR sensors (see \cref{sec:related}).
Great progress has been made in applying supervised machine learning to this problem.
However, to achieve good results, large datasets that contain LiDAR measurements with object annotations are required.
Creating such a dataset is laborious and costly.
Thus, there is only a limited number of publicly available datasets.
While some alternative datasets such as \cite{Caesar2019,Huang2018} have become available recently, most published research is based on the KITTI dataset \cite{Geiger2013}.
The limited availability of datasets also means that suitable annotated datasets are only available for a very limited set of sensors.

This paper demonstrates that a network that directly predicts objects based on data from a LiDAR sensor (in this case a Velodyne VLP-32) can be trained using data from another LiDAR sensor with significantly higher resolution (the Velodyne HDL-64E used for recording the KITTI dataset).
This is achieved by modifying the training data to make it appear like data from the lower resolution sensor.
The additional available data can be used for data augmentation.

Range images are chosen as the representation of the LiDAR measurements.
This representation is very close to the way spinning LiDAR sensors provide their measurements.
This means that no preprocessing such as assigning points to a grid or calculating features is required.
Range images implicitly contain information about occlusions where an object or parts of it are not visible due to another object closer to the sensor that is blocking the view.
Compared to common bird's eye view (BEV) or other grid based methods, this approach also allows for a relatively low runtime without imposing any constraints on the distance of detectable objects.
For CNNs that use convolutions in BEV, the invariance of convolutions under translations corresponds to translations of objects.
In contrast, range image based CNNs are only invariant under rotation around the sensor.
This may restrict their ability to generalize and may contribute to the fact that range image based networks tend to have a lower detection performance than networks using certain other representations \cite{Chen2017}.
However, \cite{Meyer2019} has shown recently that a probabilistic approach can significantly improve detection performance in networks based on range images.
To achieve this result they predict objects represented as mixture distributions, apply mean shift clustering, and use a non-maximum suppression strategy that considers uncertainties.

The structure of the neural network used in our paper is designed to be efficient enough to process a \SI{360}{\degree} range image from a Velodyne VLP-32 sensor on the experimental vehicle in real time.
Additionally, the effectiveness of a simple solution to reduce the effect of blurry predictions is demonstrated.

A discussion of related work with a focus on LiDAR-based object detection can be found in \cref{sec:related}.
\Cref{sec:design} discusses the design of the detector and the training process.
In \cref{sec:eval} the results are evaluated using both the KITTI dataset and data recorded with a Velodyne VLP-32 sensor attached to a research vehicle.

\section{Related Work} \label{sec:related}

Scanning laser sensors have been in use in research projects on self-driving vehicles for more than three decades \cite{Kanade1986}.
Early object detection was based on clustering and additional post-processing \cite{Stiller2000,Dietmayer2001}.
In recent years, there has been significant progress in object detection using LiDAR sensors due to the availability of higher resolution LiDAR sensors, publicly available datasets (especially KITTI), and the progress in deep learning.

Many of the neural networks used for detecting objects in LiDAR point clouds are based on ideas from convolutional neural networks (CNNs) that detect objects in 2D images.
This includes both methods that create a dense grid of predictions such as \cite{Redmon2016} and methods that output predictions for a previously generated set of region proposals, e.\,g.\ \cite{Ren2015}.
An overview over CNN-based 2D object detection methods can be found in \cite{Huang2017}.

To process LiDAR data in neural networks, various representations of LiDAR point clouds have been used.
One approach is to create a bird's eye view (BEV) image by projecting the LiDAR points onto a ground plane.
Information such as point density, reflectivity, and height can be encoded in channels (see e.\,g.\ \cite{Yu2017,Beltran2018}).
LiDAR data represented as range images has been used in CNNs to detect objects \cite{Li2016,Meyer2019}.
As shown in \cite{Chen2017} this representation can also be used together with other representations (including BEV) in a single network.
Their evaluation suggests that using BEV can achieve a better detection performance than methods using the range image representation.
One advantage of both BEV and range image representation is that standard 2D convolutions and network architectures that are very similar to those used in 2D object detection can be used.
This makes them relatively easy to implement and adapt.

An alternative to approaches that use a two-dimensional representation of the point cloud are network structures that operate directly on three-dimensional data.
\cite{Li2017} uses a three-dimensional grid and 3D convolutions to predict objects.
An architecture that can exploit the sparsity of data in three dimensional grids is presented in \cite{Engelcke2017}.
In \cite{Charles2017} a neural network that can operate directly on unstructured point clouds and is inherently invariant to permutations of the points in the point cloud has been proposed.
Multiple ways to adopt this idea for automotive object detection have been presented \cite{Zhou2017a,Qi2018,Lang2018}.
A comprehensive overview over various object detection methods for autonomous driving can be found in \cite{Arnold2019} and \cite{Feng2019}.

The vast majority of these object detection algorithms is evaluated using data from the type of sensor that has been used for training.
There are however some exceptions to this.
\cite{Beltran2018} shows that their BEV-based network which was trained on KITTI can also predict objects using data from lower resolution lidar sensors.
In \cite{Pino2018} a CNN for range images from a VLP-16 sensor with only 16 channels is trained using a part of the KITTI LiDAR data.
This CNN is used to perform a point-wise classification of the range image.
Then a clustering-based approach is used to extract objects.

\section{Network Architecture and Training} \label{sec:design}

This section describes the design of the proposed object detection.
In \cref{sec:io-rep,sec:model} the input and output format and structure of the neural network are discussed.
\Cref{sec:sensor-comp} describes the differences between the data from our Velodyne VLP-32 sensor and the KITTI LiDAR data.
Our approach to accommodate for these differences is presented in \cref{sec:sensor-sim}.
\Cref{sec:training,sec:postprocessing} describe the training procedure and necessary post-processing.

\subsection{Input and Output Representation} \label{sec:io-rep}

The range image (in meters) is scaled by a factor of \num{.01} (so most values are between 0 and 1) and used as input to the CNN.
For every input pixel the CNN then predicts object parameters for multiple orientation anchors.
The predicted parameters for each anchor are:
\begin{itemize}
	\item a scalar objectness score
	\item the three-dimensional offset of the box center relative to the point for which this object is predicted ($\in\mathbb R^3$) in meters
	\item the orientation expressed as a two-dimensional unit vector (this avoids discontinuities that occur when using angles to represent the orientation)
	\item the width, length, and height of the object in meters
\end{itemize}
Thus the network has to predict nine channels per anchor.
To ensure invariance under rotation around the sensor origin, the box position and object orientation are represented in a coordinate frame that is aligned with the direction of sight to the corresponding point.
The resulting box representation is effectively a three-dimensional extension of the representation in \cite{Meyer2019}.
This representation significantly reduces the number of output channel compared to the separate prediction of all corner points as found in \cite{Li2016}.

The anchors represent different object classes at different aspect angles.
For each object class there are four anchors corresponding to four ranges of possible orientations (at \SI{90}{\degree} intervals).
The estimation of the objectness score for each anchor forms a classification problem and determines the object class and approximate orientation.
Regression is used to determine the precise orientation and the other object parameters that are predicted for each anchor.
The idea to combine classification and regression to estimate orientations has been used in other object detectors such as \cite{Mousavian2017}.

\subsection{Model Design} \label{sec:model}

\begin{figure}
	\centering
	{\footnotesize 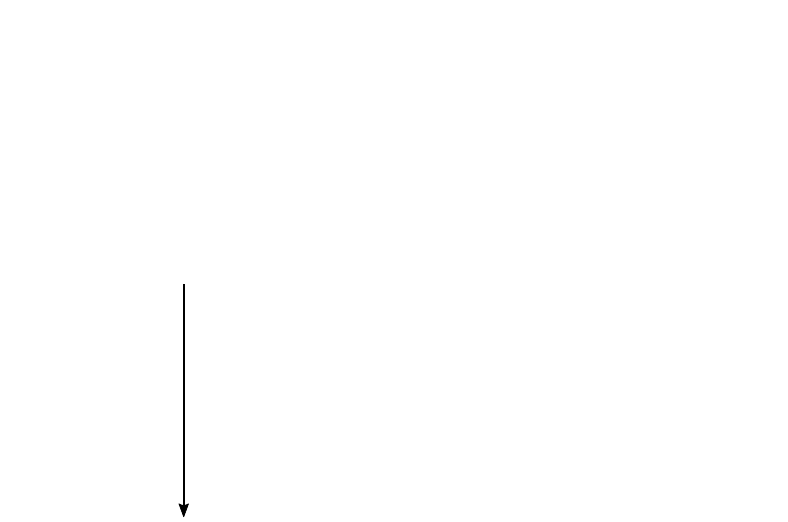 }
	\caption{Strucure of the residual network.}
	\label{fig:resnet-structure}
\end{figure}

The network is designed to be a fully convolutional network build from residual blocks \cite{He2016}.
Apart from one 1$\times$1 convolution to increase the number of channels to 64, the network is entirely build up of residual blocks.
A sketch of the network structure can be seen in \cref{fig:resnet-structure}.
Overall, it consists of 32 residual blocks, where the last one differs slightly from the intermediate blocks.
All layers use the full resolution of the range image, so no stride or pooling is required.
This reduces blurring of the predicted output.
Blurring is particularly problematic at the edges of objects in range images where neighboring pixels can represent points with significant distance to each other.

\begin{figure}
	\centering
	{\footnotesize 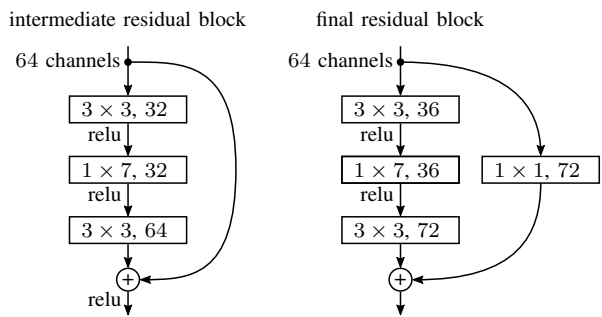 }
	\caption{Residual blocks used in the model.}
	\label{fig:resnet-block}
\end{figure}

\Cref{fig:resnet-block} depicts the structure of the individual residual blocks.
The range images have significantly more columns than rows.
Therefore, the middle layer of each convolutional block uses 1$\times$7-kernels to increase the receptive field horizontally.
Standard rectified linear units (ReLU) are used as activation functions.
As shown in \cref{fig:resnet-block}, the last residual block of the network differs slightly from the intermediate blocks.
Its number of output channels matches the shape of the desired prediction (two classes with four orientation anchors each and nine channels per anchor).
Furthermore, no activation function is applied after the last layer of the last block to allow for arbitrary predicted values.

The predicted objectness scores have a tendency to be blurry, i.\,e.\ the predicted objectness maps do not have clear edges.
This means that points that are adjacent to objects in the range image may also have high predicted objectness scores.
These inaccurate objectness scores result in wrongly detected objects that may be a lot closer or further away than the objects that should be detected.
During evaluation, each pixel in the predicted objectness score map is replaced by the minimal score in its 3$\times$5 neighborhood.
The effect of this operation is that the detections in the range image are shrunk.
This reduces the objectness around edges of objects, thereby reducing false positives.
The minimum over a neighborhood can be implemented efficiently by using an appropriately parameterized max-pooling layer.

\subsection{Sensor Comparison} \label{sec:sensor-comp}

Both the Velodyne VLP-32 sensor \cite{VelodyneLiDAR2018} mounted on the vehicle and the Velodyne HDL-64E used while recording the KITTI dataset generate a point cloud by spinning around a vertical axis while recording distance measurements with (32/64) channels.
These channels are angled to cover a range of elevation angles.
In both the recording of the KITTI dataset and our research vehicle, the sensors are configured to run at a rotational frequency of \SI{10}{\hertz}.
This sensor type allows for a representation of the point cloud as a range image, where each row corresponds to a sensor channel and each column contains measurements recorded at roughly the same time during a revolution.

\begin{figure*}
	\centering
	\includegraphics[trim=0 0.25cm 0 0.3cm,clip]{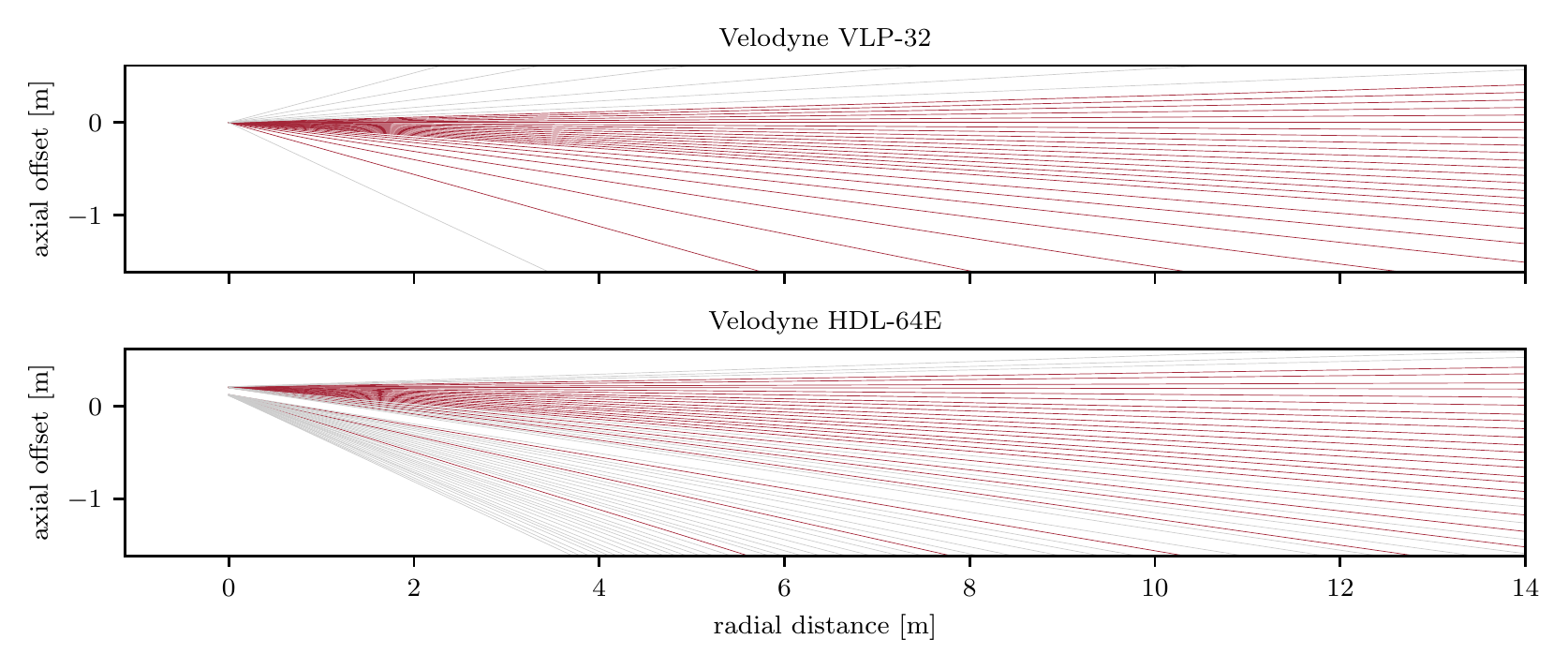}
	\caption{Comparison of the sensor channel calibration of the VLP-32 sensor and the reconstructed channel alignments from the KITTI HDL-64E data.
		In the upper plot, the 25 sensor channels used for predicting objects are highlighted in red.
		In the lower plot, one option of selecting a similar pattern from the HDL-64E data is highlighted.
		For the VLP-32 sensors, the origin is chosen to be in the optical center.
		For the HDL-64E sensor, the origin corresponds to the origin of the Velodyne coordinate frame in the KITTI dataset (located at the base of the sensor).}
	\label{fig:sensor-channels}
\end{figure*}
For the VLP-32 sensor, this results in range images with a width of approximately 1808 columns per revolution.
The HDL-64E sensor used to record the KITTI dataset has a recording frequency that results in a horizontal resolution of about 2083 columns per revolution.
The channels of the HDL-64E cover an elevation angle from \SI{-25}{\degree} (down) to \SI[retain-explicit-plus]{+2}{\degree} (up) with a resolution of about \SI{1/3}{\degree} near the horizon and a slightly lower angular resolution for the lower channels (see \cref{fig:sensor-channels}).
The VLP-32 covers an elevation angle from \SI{-25}{\degree} to \SI[retain-explicit-plus]{+15}{\degree}.
While the angular resolution around the horizon is \SI{1/3}{\degree}, the gaps between the higher and lower channels of this sensor are significantly larger.
The reflectivity measurements provided by both sensors are not considered as they differ significantly between both sensors.

\subsection{Sensor Simulation} \label{sec:sensor-sim}

The LiDAR data in the KITTI dataset is in the form of unorganized point clouds.
The sensor channel alignment is reconstructed from the point cloud data so the points (and corresponding labels) can be assigned to pixels of the range image representation.
The range images are reconstructed with the horizontal resolution of the VLP-32 sensor (1808 columns).
Only 25 rows of the range image of the VLP-32 sensor are used as input for the CNN detector.
The six highest and the lowest sensor channel are ignored, as they are not relevant for road object detection.
Some of these channels also have no corresponding data in the KITTI dataset due to the different sensor channel orientations.

The distribution and number of rows in the reconstructed range images from KITTI point clouds differ significantly from VLP-32 data.
Therefore the reconstructed range images are not well suited for training the neural network.
Instead, we want to extract range images from the KITTI data that are very similar to range images from VLP-32 sensors.
As shown in \cref{fig:sensor-channels}, a subset of channels of the HDL-64E can be chosen such that the angles between the selected channels are very similar to the angles between the channels of the VLP-32.
This yields a range image that closely resembles actual VLP-32 measurements in terms of both absolute elevation angles and angles between adjacent rows.

The entire channel selection can be moved up or down one or multiple HDL-64E channels without significantly changing the angles between the selected channels.
Furthermore, the selection of HDL-64E channels for some of the lower VLP-32 channels is somewhat ambiguous.
Combining these two degrees of freedom allows us to define 12 different channel subsets.
For each training step, one of these 12 subsets is randomly chosen and the corresponding rows from the range image and label data are used.
This randomization provides a level of data augmentation.

\subsection{Training} \label{sec:training}

The loss function consists of two components that directly compare the output of the model as depicted in \cref{fig:resnet-structure} with corresponding pixel-wise labels.
The classification loss is applied to the predicted objectness scores whereas the regression loss is only applied to the predicted box parameters.
For most object detection networks, cross-entropy or focal loss \cite{Lin2017} are used as classification losses.
As this form of loss seems to reduce detection performance, we use a quadratic loss in our work.
A mask that is applied to the classification loss makes sure that only the area in front of the vehicle where objects are labeled is used for training.
Furthermore, this mask is used to exclude points that fall into areas marked as ``don't care"-regions in the KITTI dataset.
The regression loss is calculated for each anchor separately and then summed up to obtain the overall regression loss.
For each anchor, the mean squared error with all eight regression channels weighted equally is used.
To ensure that the regression loss only considers regions labeled as objects of the corresponding anchor, a mask is applied to the regression loss of each anchor.
The overall loss is the sum of the classification and the regression loss.

The network is implemented in Tensorflow \cite{Abadi2015} and the Adam optimizer \cite{Kingma2015} is used for training the network.
During training, only the parts of the range images and labels that cover the \SI{180}{\degree} area in front of the vehicle, are used.
The initial learning rate is \num{1e-4}.
During the training process, this is reduced in steps to \num{3e-6}.

\subsection{Post-Processing} \label{sec:postprocessing}

\begin{figure}
	\centering
	\includegraphics[trim=0 0.3cm 0 1.2cm,clip]{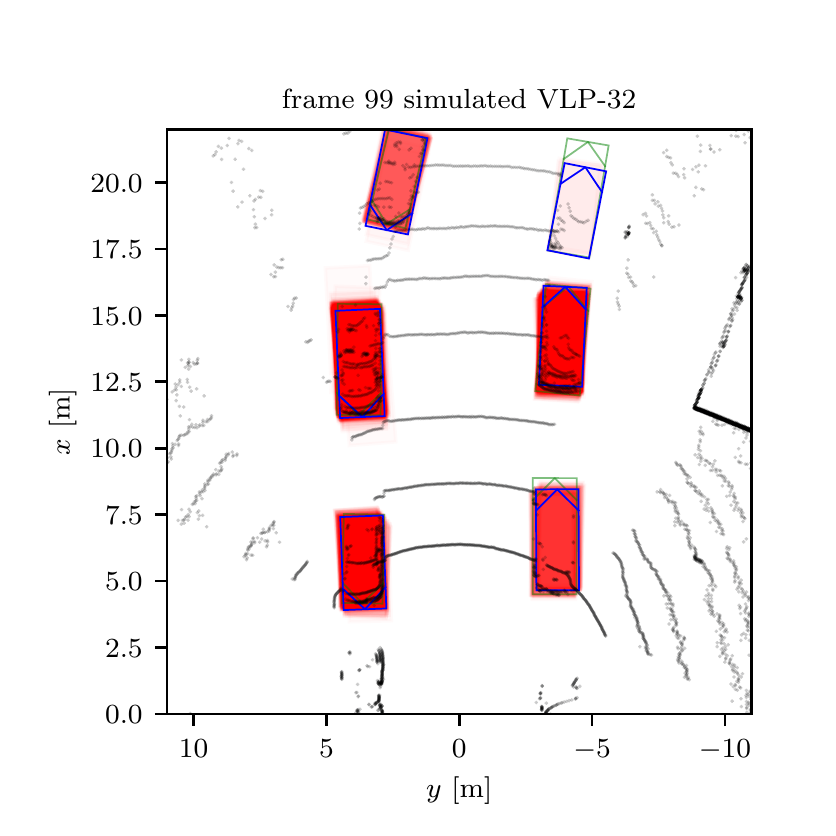}
	\caption{Detection example using a simulated VLP-32 range image based on KITTI data.
	Raw detections after applying the threshold are displayed as opaque red boxes (darker red implies more overlapping detections).
	Objects after non-maximum suppression are shown in blue.
	The labels are visible in green.}
	\label{fig:result-boxes}
\end{figure}
To reduce the impact of predictions that have a high objectness score but an inconsistent orientation vector $r$ (i.\,e.\ its norm differs significantly from 1), the objectness score is multiplied with a factor of
\begin{equation}
\min\left(\Vert r\Vert, \frac{1}{\Vert r\Vert}\right) \eqpunct{.}
\label{eq:orient-norm-factor}
\end{equation}
Ideally, the predicted $\Vert r\Vert$ should always be equal to one.
In this case, the objectness score is not affected.
Orientation vectors with a norm that is significantly larger or smaller than one indicate unreliable orientation estimates.
For these orientation vectors, \cref{eq:orient-norm-factor} reduces the corresponding objectness score.
Then a threshold is applied to the predicted objectness scores.

There are often thousands of predicted objects remaining after thresholding.
To limit the number of detections and avoid overlapping objects non-maximum suppression (NMS) is used.
It uses a BEV grid representation where each cell contains information about the object with the highest score that covers the cell.
All object detections are ordered in descending order by score.
Then the algorithm iterates over all detections.
Objects are marked invalid if the covered cells are already occupied by a valid object.
Otherwise, a reference to the objects is stored in all cells they cover.
The remaining valid objects form a set of detections that do not overlap each other.
\Cref{fig:result-boxes} shows an example of the detection before and after applying the non-maximum suppression.
This NMS algorithm is implemented in C++ and runs on the CPU.
For the following evaluation, a grid size of \SI{.2}{\meter} is used.

\section{Evaluation} \label{sec:eval}

We have split the official \emph{training} data from the KITTI object detection dataset into two subsets, one for training and one for validation.
To reduce similarity between training and validation data, all frames that are in the same sequence are also grouped into the same subset.
Overall, this leaves 6090 of the 7481 frames in the \emph{training} dataset for training and 1391 for validation.

For evaluation the network is trained with two classes.
One class consists of the \emph{car} and \emph{van} class in the KITTI dataset.
The other one is a combination of the \emph{pedestrian}, \emph{cyclist}, and \emph{person sitting} classes.
(The KITTI evaluation is only available for \emph{car} and \emph{pedestrian}.)
Due to a strong bias of labels in the KITTI dataset towards cars and the planned usage of the detector primarily for vehicle detection, the focus of the evaluation is on the detection of cars.

\subsection{Evaluation on the KITTI Dataset}

The evaluation code provided by KITTI for the object detection dataset was used for evaluations on the validation dataset.
The KITTI evaluation code allows for duplicate detections for a single label.
Therefore generating multiple overlapping detections may be beneficial to achieve a higher recall.
However, the non-maximum suppression strategy used (and required for additional processing steps on the experimental vehicle) cannot benefit from this.
Using the evaluation code with the raw predictions is also impractical, as the number of those predictions is too large.
Vans detected as cars and sitting persons detected as pedestrians are not considered errors by the evaluation code.
The KITTI object detection benchmark provides three different evaluations.
One considers the intersection over union (IoU) of 2D object detections and the corresponding labels in the image.
We generate these bounding boxes by projecting the 3D boxes into the image.
Additionally there is a bird's-eye view evaluation that considers the IoU of detections and labels in the bird's-eye view and a three-dimensional evaluation that evaluates the three-dimensional IoU of the boxes.
For an object to be considered detected, the two-dimensional/three-dimensional IoU needs to be greater than \SI{70}{\percent} for cars and greater than \SI{50}{\percent} for other classes.
The KITTI dataset groups objects into the difficulty levels ``easy", ``moderate" and ``hard" depending on the occlusion level and 2D bounding box height.

\begin{figure}
	\centering
	\includegraphics{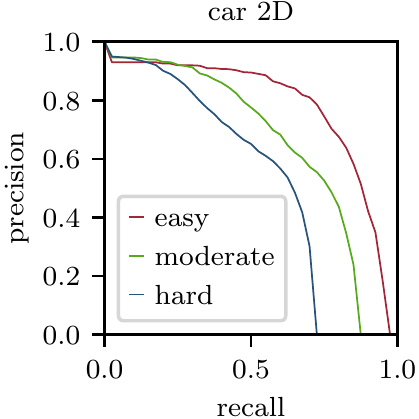}
	\includegraphics{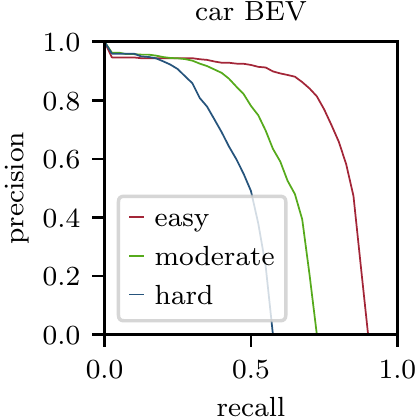}
	\caption{Precision recall curves of the 2D (left) and BEV (right) evaluation.
	}
	\label{fig:pr}
\end{figure}

For the evaluation on the KITTI dataset, a fixed channel subset of the validation data is used to provide deterministic results.
Objects are extracted with a low objectness threshold of \num{.05}.
This allows for a wide range of thresholds when computing precision-recall curves.
\Cref{fig:pr} shows the precision recall curves for the projection to the image and the BEV.
The difference between these curves suggests that many objects are detected, while their box is not accurate enough to pass the \SI{70}{\percent} threshold in BEV.
Additionally, the recall drops significantly for harder labels.
In \cref{tab:evals} a comparison between the average precision (AP) values of the different evaluations can be seen.
\begin{table}
	\caption{Performance in different evaluations}
	\label{tab:evals}
	\centering
	\begin{tabular}{lccc}
		\toprule
		& \multicolumn{3}{c}{vehicle AP} \\
		\cmidrule(lr){2-4}
		& easy & moderate & hard \\
		\midrule
		2D object & \SI{77.53}{\percent} & \SI{65.54}{\percent} & \SI{53.20}{\percent} \\
		BEV & \SI{75.89}{\percent} & \SI{57.71}{\percent} & \SI{44.13}{\percent} \\
		3D object & \SI{52.98}{\percent} & \SI{38.86}{\percent} & \SI{28.24}{\percent} \\
		\bottomrule
	\end{tabular}
\end{table}
A comparison to the reported average precision for the 2D evaluation on the testing dataset in \cite{Li2016} (\SI{60.3}{\percent} / \SI{47.5}{\percent} / \SI{42.7}{\percent}) suggests a significant improvement.
Compared to \cite{Meyer2019}, the BEV results are only slightly worse for the easy category (\SI{75.89}{\percent} vs. \SI{78.25}{\percent}).
However the difference increases towards the ``hard" category (\SI{44.13}{\percent} vs. \SI{66.47}{\percent}).
This is probably related to the significantly more sophisticated (and time consuming) non-maximum suppression strategy used in their approach.
State-of-the-art algorithms using more complex point cloud representations (such as \cite{Lang2018}) achieve significantly higher average precision values.

\Cref{tab:ablation} shows the effect of certain components of the object detector and the training strategy on the detection performance.
\begin{table}
	\caption{Ablation Study KITTI BEV}
	\label{tab:ablation}
	\centering
	\begin{tabular}{lccc}
		\toprule
		& \multicolumn{3}{c}{vehicle AP} \\
		\cmidrule(lr){2-4}
		& easy & moderate & hard \\
		\midrule
		ours & \SI{75.89}{\percent} & \SI{57.71}{\percent} & \SI{41.13}{\percent} \\
		no neighbor minimum & \SI{64.22}{\percent} & \SI{51.90}{\percent} & \SI{39.84}{\percent} \\
%		no 1$\times$7 convolutions & & & \\
		no orientation achors & \SI{45.20}{\percent} & \SI{37.96}{\percent} & \SI{30.16}{\percent} \\
		fixed training channels & \SI{69.45}{\percent} & \SI{55.74}{\percent} & \SI{41.98}{\percent} \\
		\bottomrule
	\end{tabular}
\end{table}
If the layer that calculates the minimum in a neighborhood of pixels in the predicted objectness scores is not applied, the average precision for the easy and moderate categories drops, while the hard category is affected less.
This is somewhat expected, as this layer may not only remove wrongly predicted objects from pixels around the edges of objects in the range image but also reduce the objectness score of distant, small, or partially visible objects.
These might be handled better by approaches similar to the mean shift clustering used in \cite{Meyer2019} (at the cost of complexity and runtime).
If no orientation anchors are used and only a single set of regression channels is predicted, the detection performance drops significantly.
This might be due to situations where the orientation is somewhat ambiguous and the predictions of the network without orientation anchors effectively predict a combination of box parameters for multiple possible orientations.
Using a fixed subset set of training channels instead of randomly switching between twelve sets of channels during training reduces the average precision by a few percentage points.

\subsection{Evaluation Using Data from VLP-32 Sensor}

\begin{figure}
	\centering
	\includegraphics[width=\linewidth,trim=0cm 1.5cm 0cm 0cm,clip]{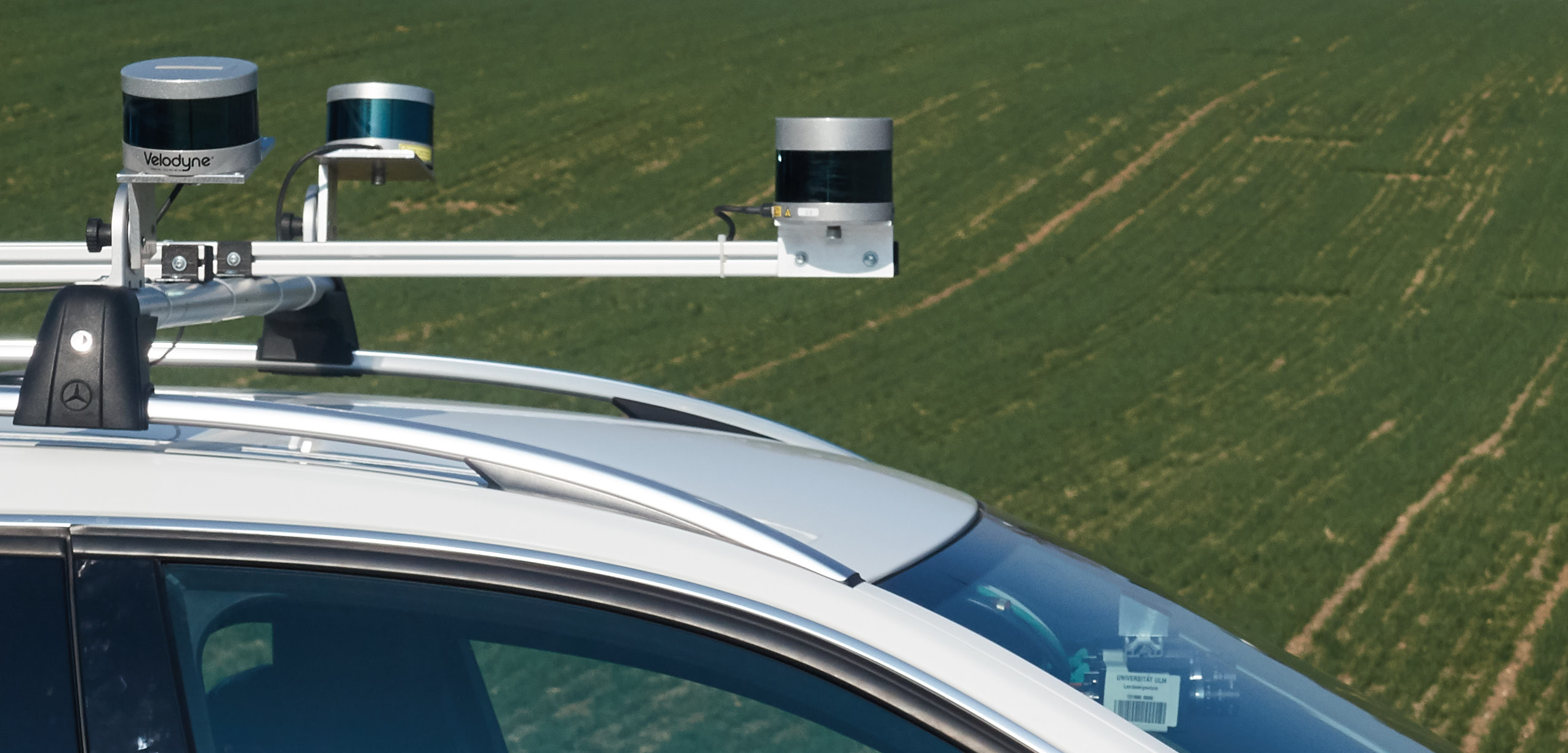}
	\caption{Velodyne VLP-32 mounted on the experimental vehicle.
		Only the Velodyne VLP-32 in the front was used for this paper.
	}
	\label{fig:vehicle}
\end{figure}

Our experimental vehicle is equipped with a variety of sensors including a Velodyne VLP-32 that is mounted over the front of the vehicle's roof (see \cref{fig:vehicle}).
This mounting position results in a similar height of the sensor over ground as the Velodyne HDL-64E used for recording the KITTI dataset.

There is no appropriately labeled data from the research vehicle available.
\begin{figure}
	\centering
	% 08:01:37.184 in Rec20180824095950_U1800 (about 106.6s into recording)
	\resizebox{\linewidth}{!}{
		\includegraphics[height=.5\linewidth,trim=5cm 5cm 36cm 11cm,clip]{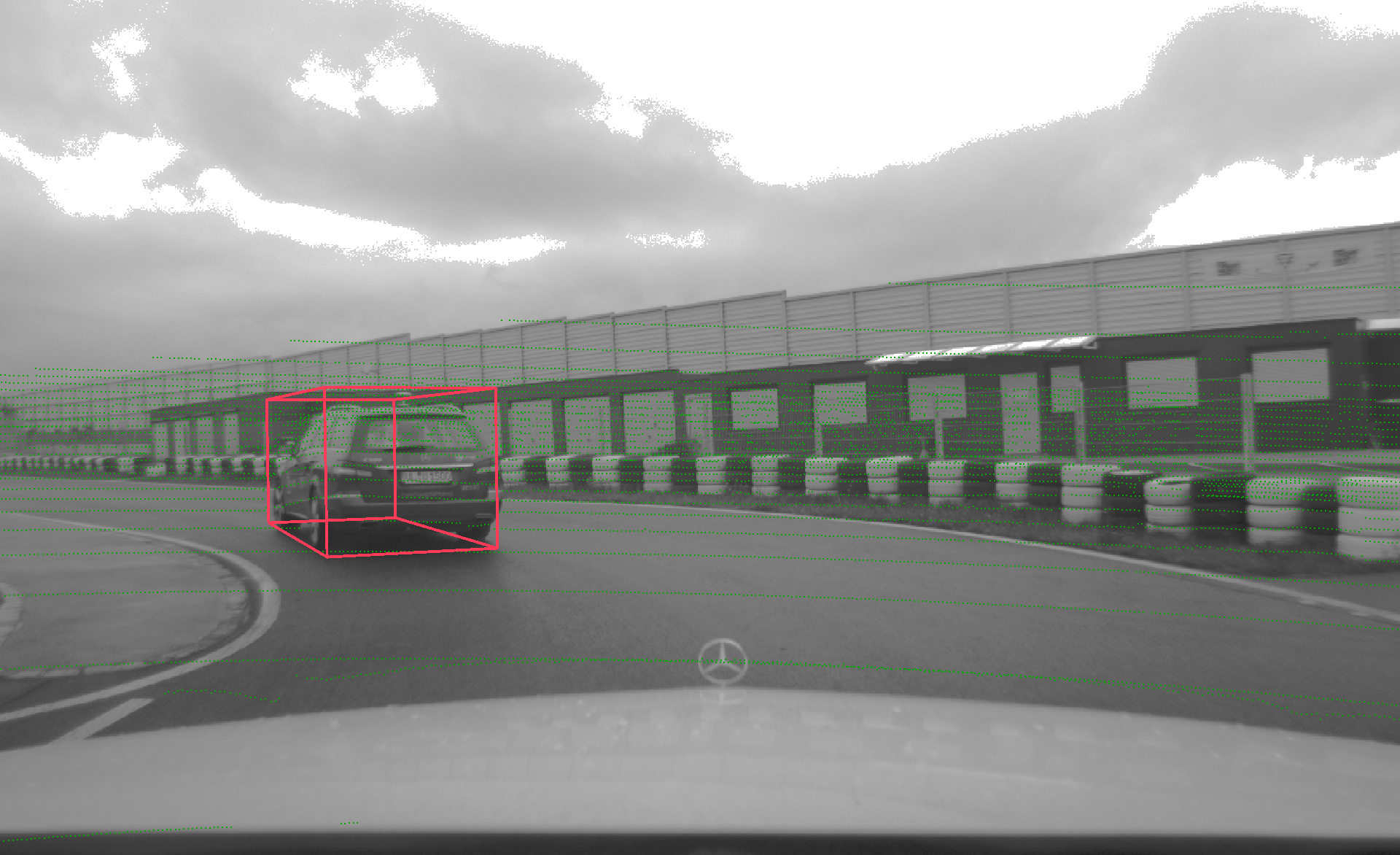}
		\includegraphics[height=.5\linewidth,trim=20cm 7.5cm 12cm 7cm,clip]{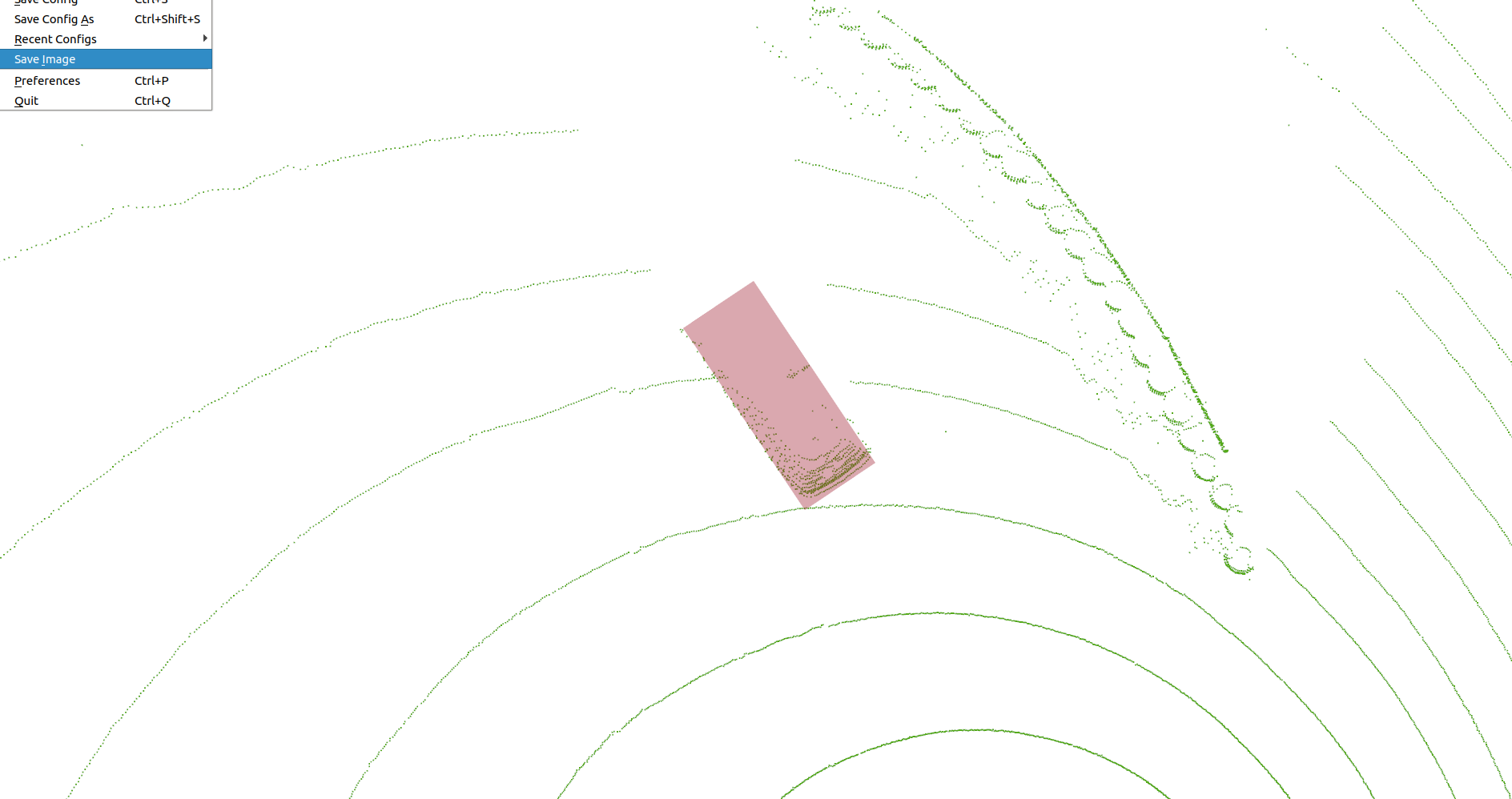}
	}
	\caption{Detected reference vehicle in an evaluation sequence.}
	\label{fig:ref-vehicle}
\end{figure}
To evaluate the performance on the research vehicle, data was recorded while driving together with a reference vehicle (see \cref{fig:ref-vehicle}).
Both vehicles are equipped with DGPS and inertial measurement units to accurately determine their own position.
The recorded position of the reference vehicle can be used as ground truth data for evaluating the accuracy of the detection.
\Cref{tab:refcar-eval} shows the detection ratio of the reference vehicle and the root-mean-square error (RMSE) of the detection depending on the distance of the reference vehicle from the vehicle with the LiDAR sensor.
\begin{table}
	\caption{Evaluation with Reference Vehicle}
	\label{tab:refcar-eval}
	\centering
	\begin{tabular}{lccc}
		\toprule
		& \multicolumn{3}{c}{ref.\ vehicle at distance (in \si{\meter})} \\
		\cmidrule(lr){2-4}
		& 0 -- 20 & 20 -- 40 & $>$ 40 \\
		\midrule
		evaluation frames & 663 & 1053 & 293 \\
		detection ratio & \SI{97.0}{\percent} & \SI{80.9}{\percent} & \SI{25.3}{\percent} \\
		radial position error (RMSE) [\si{\meter}] & 0.41 & 0.58 & 0.67 \\
		tangential position error (RMSE) [\si{\meter}] & 0.15 & 0.24 & 0.32 \\
		vertical position error (RMSE) [\si{\meter}] & 0.13 & 0.26 & 0.59 \\
		orientation error (RMSE) [\si{\degree}] & 2.3 & 7.3 & 16.4 \\
		length error (RMSE) [\si{\meter}] & 0.88 & 1.13 & 1.22 \\
		width error (RMSE) [\si{\meter}] & 0.19 & 0.22 & 0.23 \\
		height error (RMSE) [\si{\meter}] & 0.08 & 0.07 & 0.07 \\
		\bottomrule
	\end{tabular}
\end{table}

When the reference vehicle is close, it is detected very reliably.
As the distance increases, the detection ratio drops significantly.
The position error is split into its radial component, i.\,e.\ the error in estimating the distance to the object's center, its tangential component that describes the position error to the side of the viewing direction, and the vertical error.
Overall, position error, orientation error, and box size estimation error increase with larger distance.
Presumably this is caused by a smaller number of labeled objects at larger distances in the training data.
Furthermore, these distant objects form relatively small patches in range images.
Orientation estimation and classification seem to be most affected by the limited number of points that belong to each of these distant objects.

The position error in radial direction and the error of the length estimate are larger than position and size estimates along other axes.
This may be caused by frames in the recorded sequence where only the rear of the reference vehicle is visible and the vehicle's length and consequently the distance to its center is underestimated.

When processing full \SI{360}{\degree} range images from the Velodyne VLP-32 sensor spinning at 10 revolutions per second, the average inference time on a computer with a Nvidia GeForce 1080~Ti GPU is \SI{39.8}{\milli\second}.
The average duration of the non-maximum suppression is \SI{0.251}{\milli\second}.
This means that our object detector can be used in real time on the research vehicle.

Qualitative analysis of the performance in urban scenarios shows that close-by objects on the side of the vehicle are not detected reliably.
As the KITTI dataset only provides labels in the area in front of the vehicle that is visible in the camera image, this may be a limitation of the dataset.
However, for objects at distances that are more typical for labeled objects in the KITTI dataset the object detector can generalize to objects on the side of the car or behind it.

\section{Conclusion} \label{sec:conclusion}

This paper shows that labeled data from one type of LiDAR sensor can be used for training a neural network that can provide object detections from data generated by a different type of sensor.
It also shows that relatively simple improvements to the network can significantly improve the performance of object detection from range images without a significant impact on runtime performance.
This points to the fact that the type of CNN employed here struggles to provide accurate predictions around the edges of objects in range images.
The presented object detection approach can be used in real time for detecting objects around the vehicle.
However, there is still a significant difference in detection performance of neural networks that use range images compared to the detectors that currently rank highest in the KITTI benchmark.
By combining the results from this paper with the ideas recently presented in \cite{Meyer2019} to improve object detection from range images it might be possible to further reduce this performance gap.
Our training strategy could also allow for combining labeled data from multiple types of sensors for training a neural network.

\bibliographystyle{IEEEtran}

\bibliography{IEEEabrv,references.bib}

% \cleardoublepage

% \section{Supplementary Materials}

\end{document}

%% file: img/resnet-layout.pdf_tex
%% Accompanies image file 'resnet-layout.pdf' (pdf, eps, ps)
\begingroup%
  \makeatletter%
  \providecommand\color[2][]{%
    \errmessage{(Inkscape) Color is used for the text in Inkscape, but the package 'color.sty' is not loaded}%
    \renewcommand\color[2][]{}%
  }%
  \providecommand\transparent[1]{%
    \errmessage{(Inkscape) Transparency is used (non-zero) for the text in Inkscape, but the package 'transparent.sty' is not loaded}%
    \renewcommand\transparent[1]{}%
  }%
  \providecommand\rotatebox[2]{#2}%
  \newcommand*\fsize{\dimexpr\f@size pt\relax}%
  \newcommand*\lineheight[1]{\fontsize{\fsize}{#1\fsize}\selectfont}%
  \ifx\svgwidth\undefined%
    \setlength{\unitlength}{226.77165354bp}%
    \ifx\svgscale\undefined%
      \relax%
    \else%
      \setlength{\unitlength}{\unitlength * \real{\svgscale}}%
    \fi%
  \else%
    \setlength{\unitlength}{\svgwidth}%
  \fi%
  \global\let\svgwidth\undefined%
  \global\let\svgscale\undefined%
  \makeatother%
  \begin{picture}(1,0.6625)%
    \lineheight{1}%
    \setlength\tabcolsep{0pt}%
    \put(0,0){\includegraphics[width=\unitlength,page=1]{resnet-layout.pdf}}%
    \put(0.2274416,0.53595543){\color[rgb]{0,0,0}\makebox(0,0)[t]{\lineheight{1.25}\smash{\begin{tabular}[t]{c}$1\times 1$, $64$\end{tabular}}}}%
    \put(0,0){\includegraphics[width=\unitlength,page=2]{resnet-layout.pdf}}%
    \put(0.22109165,0.61777153){\color[rgb]{0,0,0}\makebox(0,0)[rt]{\lineheight{1.25}\smash{\begin{tabular}[t]{r}range image\end{tabular}}}}%
    \put(0.40463228,0.31204338){\color[rgb]{0,0,0}\makebox(0,0)[lt]{\lineheight{1.25}\smash{\begin{tabular}[t]{l}31 intermediate residual blocks\end{tabular}}}}%
    \put(0,0){\includegraphics[width=\unitlength,page=3]{resnet-layout.pdf}}%
    \put(0.40463228,0.07478752){\color[rgb]{0,0,0}\makebox(0,0)[lt]{\lineheight{1.25}\smash{\begin{tabular}[t]{l}final residual block\end{tabular}}}}%
    \put(0.40463228,0.53595543){\color[rgb]{0,0,0}\makebox(0,0)[lt]{\lineheight{1.25}\smash{\begin{tabular}[t]{l}input layer\end{tabular}}}}%
    \put(0,0){\includegraphics[width=\unitlength,page=4]{resnet-layout.pdf}}%
  \end{picture}%
\endgroup%

%% file: img/resnet-block.pdf_tex
%% Accompanies image file 'resnet-block.pdf' (pdf, eps, ps)
\begingroup%
  \makeatletter%
  \providecommand\color[2][]{%
    \errmessage{(Inkscape) Color is used for the text in Inkscape, but the package 'color.sty' is not loaded}%
    \renewcommand\color[2][]{}%
  }%
  \providecommand\transparent[1]{%
    \errmessage{(Inkscape) Transparency is used (non-zero) for the text in Inkscape, but the package 'transparent.sty' is not loaded}%
    \renewcommand\transparent[1]{}%
  }%
  \providecommand\rotatebox[2]{#2}%
  \newcommand*\fsize{\dimexpr\f@size pt\relax}%
  \newcommand*\lineheight[1]{\fontsize{\fsize}{#1\fsize}\selectfont}%
  \ifx\svgwidth\undefined%
    \setlength{\unitlength}{226.77165354bp}%
    \ifx\svgscale\undefined%
      \relax%
    \else%
      \setlength{\unitlength}{\unitlength * \real{\svgscale}}%
    \fi%
  \else%
    \setlength{\unitlength}{\svgwidth}%
  \fi%
  \global\let\svgwidth\undefined%
  \global\let\svgscale\undefined%
  \makeatother%
  \begin{picture}(1,0.525)%
    \lineheight{1}%
    \setlength\tabcolsep{0pt}%
    \put(0,0){\includegraphics[width=\unitlength,page=1]{resnet-block.pdf}}%
    \put(0.20438397,0.33807373){\color[rgb]{0,0,0}\makebox(0,0)[t]{\lineheight{1.25}\smash{\begin{tabular}[t]{c}$3\times 3$, $32$\end{tabular}}}}%
    \put(0,0){\includegraphics[width=\unitlength,page=2]{resnet-block.pdf}}%
    \put(0.20438397,0.23807373){\color[rgb]{0,0,0}\makebox(0,0)[t]{\lineheight{1.25}\smash{\begin{tabular}[t]{c}$1\times 7$, $32$\end{tabular}}}}%
    \put(0,0){\includegraphics[width=\unitlength,page=3]{resnet-block.pdf}}%
    \put(0.20438397,0.13807373){\color[rgb]{0,0,0}\makebox(0,0)[t]{\lineheight{1.25}\smash{\begin{tabular}[t]{c}$3\times 3$, $64$\end{tabular}}}}%
    \put(0,0){\includegraphics[width=\unitlength,page=4]{resnet-block.pdf}}%
    \put(0.21026475,0.05769463){\color[rgb]{0,0,0}\makebox(0,0)[t]{\lineheight{1.25}\smash{\begin{tabular}[t]{c}+\end{tabular}}}}%
    \put(0,0){\includegraphics[width=\unitlength,page=5]{resnet-block.pdf}}%
    \put(0.19934534,0.02391853){\color[rgb]{0,0,0}\makebox(0,0)[rt]{\lineheight{1.25}\smash{\begin{tabular}[t]{r}relu\end{tabular}}}}%
    \put(0.19934534,0.29710236){\color[rgb]{0,0,0}\makebox(0,0)[rt]{\lineheight{1.25}\smash{\begin{tabular}[t]{r}relu\end{tabular}}}}%
    \put(0.19934534,0.19698105){\color[rgb]{0,0,0}\makebox(0,0)[rt]{\lineheight{1.25}\smash{\begin{tabular}[t]{r}relu\end{tabular}}}}%
    \put(0,0){\includegraphics[width=\unitlength,page=6]{resnet-block.pdf}}%
    \put(0.19803404,0.41988983){\color[rgb]{0,0,0}\makebox(0,0)[rt]{\lineheight{1.25}\smash{\begin{tabular}[t]{r}$64$ channels\end{tabular}}}}%
    \put(0,0){\includegraphics[width=\unitlength,page=7]{resnet-block.pdf}}%
    \put(0.65248613,0.33807373){\color[rgb]{0,0,0}\makebox(0,0)[t]{\lineheight{1.25}\smash{\begin{tabular}[t]{c}$3\times 3$, $36$\end{tabular}}}}%
    \put(0,0){\includegraphics[width=\unitlength,page=8]{resnet-block.pdf}}%
    \put(0.65248613,0.23807373){\color[rgb]{0,0,0}\makebox(0,0)[t]{\lineheight{1.25}\smash{\begin{tabular}[t]{c}$1\times 7$, $36$\end{tabular}}}}%
    \put(0,0){\includegraphics[width=\unitlength,page=9]{resnet-block.pdf}}%
    \put(0.65248613,0.13807373){\color[rgb]{0,0,0}\makebox(0,0)[t]{\lineheight{1.25}\smash{\begin{tabular}[t]{c}$3\times 3$, $72$\end{tabular}}}}%
    \put(0,0){\includegraphics[width=\unitlength,page=10]{resnet-block.pdf}}%
    \put(0.65836697,0.05769463){\color[rgb]{0,0,0}\makebox(0,0)[t]{\lineheight{1.25}\smash{\begin{tabular}[t]{c}+\end{tabular}}}}%
    \put(0,0){\includegraphics[width=\unitlength,page=11]{resnet-block.pdf}}%
    \put(0.64744749,0.29710236){\color[rgb]{0,0,0}\makebox(0,0)[rt]{\lineheight{1.25}\smash{\begin{tabular}[t]{r}relu\end{tabular}}}}%
    \put(0.64744749,0.19698105){\color[rgb]{0,0,0}\makebox(0,0)[rt]{\lineheight{1.25}\smash{\begin{tabular}[t]{r}relu\end{tabular}}}}%
    \put(0,0){\includegraphics[width=\unitlength,page=12]{resnet-block.pdf}}%
    \put(0.64613619,0.41988983){\color[rgb]{0,0,0}\makebox(0,0)[rt]{\lineheight{1.25}\smash{\begin{tabular}[t]{r}$64$ channels\end{tabular}}}}%
    \put(0,0){\includegraphics[width=\unitlength,page=13]{resnet-block.pdf}}%
    \put(0.88308954,0.23807373){\color[rgb]{0,0,0}\makebox(0,0)[t]{\lineheight{1.25}\smash{\begin{tabular}[t]{c}$1\times 1$, $72$\end{tabular}}}}%
    \put(0,0){\includegraphics[width=\unitlength,page=14]{resnet-block.pdf}}%
    \put(0.20974557,0.48781204){\color[rgb]{0,0,0}\makebox(0,0)[t]{\lineheight{1.25}\smash{\begin{tabular}[t]{c}intermediate residual block\end{tabular}}}}%
    \put(0.65784769,0.48781204){\color[rgb]{0,0,0}\makebox(0,0)[t]{\lineheight{1.25}\smash{\begin{tabular}[t]{c}final residual block\end{tabular}}}}%
  \end{picture}%
\endgroup%